\pdfoutput=1

\documentclass[11pt]{article}

\usepackage[final]{acl}

\usepackage{times}
\usepackage{tabularx}
\usepackage{latexsym}
\usepackage{amsmath}
\usepackage{color, colortbl} 

\usepackage[T1]{fontenc}

\usepackage[utf8]{inputenc}

\usepackage{microtype}

\usepackage{inconsolata}
\usepackage{multirow}

\usepackage{graphicx}
\usepackage{booktabs}
\usepackage{amssymb}
\usepackage{pifont}
\newcommand{\cmark}{\ding{51}}%
\usepackage{authblk}

\title{LegoSLM: Connecting LLM with Speech Encoder using CTC Posteriors}

\author{
Rao Ma\textsuperscript{1}\thanks{Work done during internship at Google.}, Tongzhou Chen\textsuperscript{2}, Kartik Audhkhasi\textsuperscript{2}, Bhuvana Ramabhadran\textsuperscript{2} \\
\textsuperscript{1}University of Cambridge \quad
\textsuperscript{2}Google Deepmind
}

\definecolor{Gray}{gray}{0.9}

\begin{document}
\maketitle
\begin{abstract}
Recently, large-scale pre-trained speech encoders and Large Language Models (LLMs) have been released, which show state-of-the-art performance on a range of spoken language processing tasks including Automatic Speech Recognition (ASR). To effectively combine both models for better performance, continuous speech prompts, and ASR error correction have been adopted. However, these methods are prone to suboptimal performance or are inflexible. In this paper, we propose a new paradigm, LegoSLM, that bridges speech encoders and LLMs using the ASR posterior matrices. The speech encoder is trained to generate Connectionist Temporal Classification (CTC) posteriors over the LLM vocabulary, which are used to reconstruct pseudo-audio embeddings by computing a weighted sum of the LLM input embeddings. These embeddings are concatenated with text embeddings in the LLM input space. Using the well-performing USM and Gemma models as an example, we demonstrate that our proposed LegoSLM method yields good performance on both ASR and speech translation tasks. By connecting USM with Gemma models, we can get an average of 49\% WERR over the USM-CTC baseline on 8 MLS testsets. The trained model also exhibits modularity in a range of settings -- after fine-tuning the Gemma model weights, the speech encoder can be switched and combined with the LLM in a zero-shot fashion. Additionally, we propose to control the decode-time influence of the USM and LLM using a softmax temperature, which shows effectiveness in domain adaptation.
\end{abstract}

\section{Introduction}


With the advancement of self-supervised and semi-supervised learning, large-scale pre-trained speech and text models have been released in recent years \cite{bommasani2021opportunities}. 
Today, speech encoders are pre-trained on extensive datasets that cover a wide range of spoken languages \cite{barrault2023seamlessm4t, zhang2023google, pratap2024scaling}. 
These models have achieved state-of-the-art performance in various spoken language processing tasks, including automatic speech recognition (ASR) and automatic speech translation (AST).
In the field of natural language processing (NLP), large language models (LLMs) aim to capture general world knowledge in the network parameters through the task of next-word prediction \cite{touvron2023llama, achiam2023gpt}. After being pre-trained on vast text corpora, LLMs have demonstrated remarkable capabilities in complex language understanding tasks
facilitated by prompt engineering.

\begin{figure}[!htbp]
    \centering
    \includegraphics[width=0.9\linewidth]{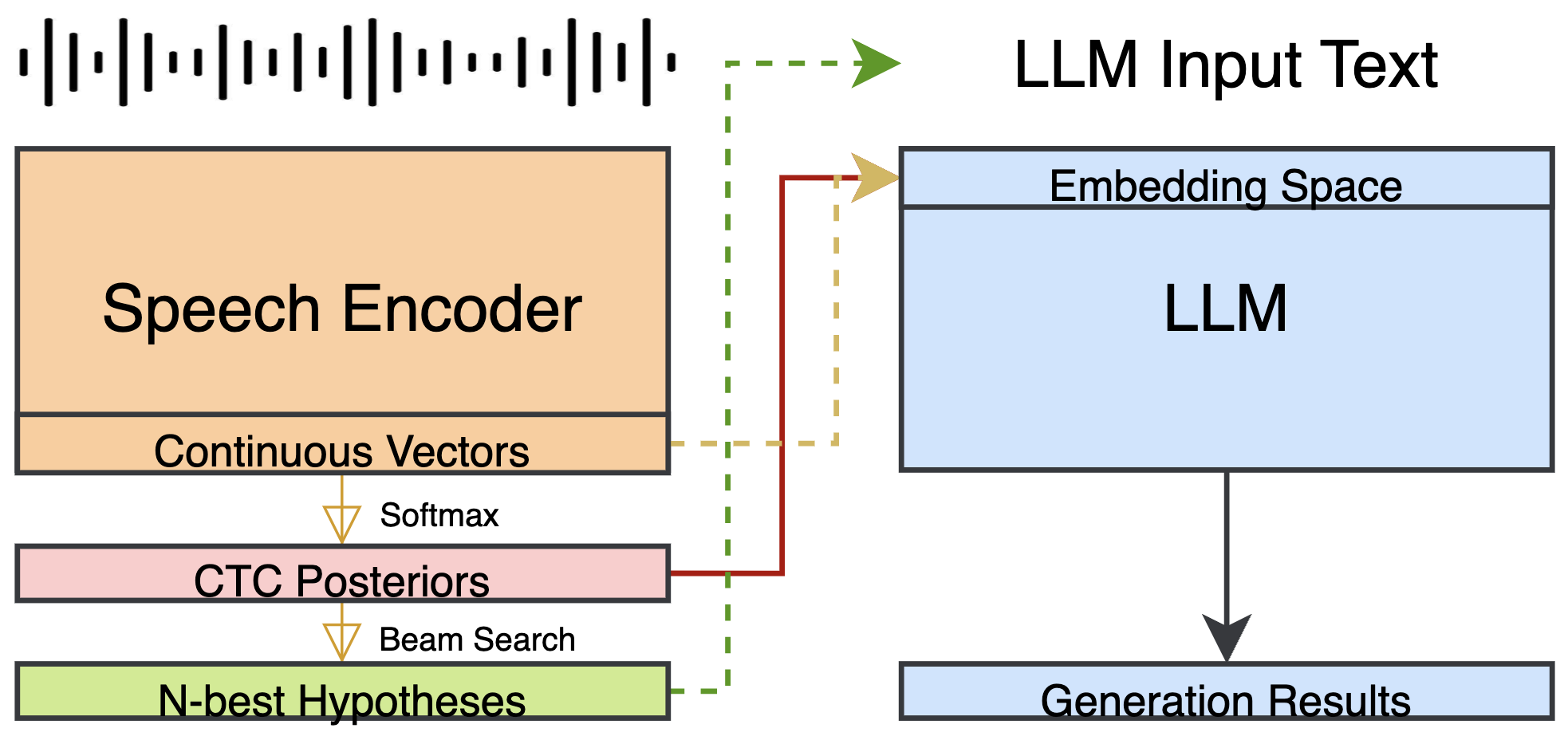}    
    \caption{Comparison of different connection methods: {ASR error correction (in green)}, {speech prompts (in orange)}, and the proposed {LegoSLM (in red)}.}
    \label{fig:scheme}
\end{figure}

To enhance the performance of spoken language processing tasks, several studies have focused on integrating speech encoders with LLMs. In ASR error correction (AEC), a cascaded system is built where the decoded hypotheses from the ASR system are given as input to the LLMs for correction \cite{errattahi2018automatic}. Such a method does not require deep access to the ASR system and has the advantage of being modular. Previous work has shown that an LLM trained on the outputs from one ASR encoder can be re-used to correct the outputs of other speech encoders without any retraining \cite{ma2023adapting}. However, the AEC performance is constrained by the limited contextual information accessible to LLM \cite{zhu2021improving}. 

Another popular approach to equipping LLMs with speech processing ability is to prompt LLMs with vectors transformed from the speech encoder output \cite{verdini2024connect}. Usually, a mapping network is inserted and trained to match the embedding space of the speech and text modalities \cite{fathullah2024prompting, gaido-etal-2024-speech}. 
By directly passing the continuous encoder outputs, this approach largely mitigates the information loss encountered in cascaded AEC systems.
While speech prompts demonstrate strong performance on various datasets, this approach sacrifices some flexibility. 
The LLM is bonded with a specific speech encoder and fails to perform the task when prompted with outputs from a different speech encoder.







In this paper, we present LegoSLM, a new paradigm to bridge pre-trained speech encoders and LLMs, as shown with the red line in Figure \ref{fig:scheme}. First, the pre-trained speech encoder is fine-tuned with Connectionist Temporal Classification (CTC) loss using the same vocabulary as the LLM. Afterward, we multiply output CTC posteriors with the LLM embedding table to reproduce pseudo-speech embeddings as the LLM input.
Compared to traditional AEC methods where the ASR hypotheses are decoded and passed, much more information is preserved with our proposed approach. At the same time, our approach maintains flexibility.
The CTC posteriors, as opposed to the continuous ASR outputs, are acquired, which helps to protect the speaker's privacy.
Moreover, after fine-tuning the LLM weights, new speech encoders can be plugged in a zero-shot fashion.
Using the USM \cite{zhang2023google} and Gemma \cite{team2024gemma} models as an example, we design comprehensive experiments to study the system performance on ASR and AST tasks. Experiments on ASR demonstrate that LegoSLM outperforms AEC and achieves performance comparable to the speech prompt method where the encoder weights are kept frozen. 
On AST, the proposed LegoSLM model shows improved performance over all baseline systems.



\section{Related Work}
\paragraph{Modular ASR Approach}
Within the domain of E2E ASR systems, \cite{dalmia2023legonn, botros2023lego} focus on building modular architectures. Similar to our proposed method, their methods train a decoder network that embeds the CTC posteriors generated by the speech encoder and outputs refined ASR hypotheses. This design allows encoders and decoders trained in different setups to be seamlessly combined. 
In contrast to these approaches, our work employs a decoder-only Transformer model instead of using an encoder-decoder architecture. 
Notably, our method connects a pre-trained speech encoder with LLM, rather than training the decoder weights from scratch. By leveraging the extensive world knowledge acquired during the LLM's pre-training, 
we aim to enhance the overall system performance. 
Furthermore, we introduce several novel extensions of LegoSLM.

\paragraph{ASR Error Correction} AEC is a widely used post-processing approach to enhance the overall performance of an ASR system. It takes the ASR hypotheses as input and is trained on the reference text to automatically detect and correct recognition errors \cite{mani2020asr}. Since it only requires decoding hypotheses from the ASR system, AEC can be applied to API-based services without requiring in-depth access \cite{ma2023adapting}. Several studies leverage ASR N-best lists as input instead of using the top-1 hypothesis, as these provide richer information and have been shown to enhance the system performance \cite{zhu2021improving, ma2023nbest}. Recent works build AEC models using powerful LLMs to leverage their superior language understanding and reasoning capabilities \cite{ma2023can, chen2024hyporadise, li2024investigating}. 

\paragraph{LLM with Speech Prompts}
The success of LLMs in text processing has driven their application to other modalities, such as vision \cite{wang2024visionllm} and speech \cite{tangsalmonn}. In the speech domain, a mapping network can be used to bridge speech encoders and LLMs,  which transform the outputs from the speech encoder into acoustic prompts that are compatible with the LLM text embedding space \cite{verdini2024connect, hono2023integration}. 
Various designs for the mapping network exist, and even a basic projection layer has demonstrated strong performance in aligning both modalities \cite{fathullah2024prompting, ma2024embarrassingly}. As outputs from the speech encoder can be quite long, some approaches propose to reduce the sequence length by stacking multiple vectors or employing a CTC-based compressor \cite{wu2023decoder}.







\section{Methodology}


\subsection{Structure of Speech Encoder} 
Given input audio features $\mathbf{X}_{1:T}$, the pre-trained speech encoder transform them into a sequence of hidden representations $\mathbf{h}_{1:T'}$,
\begin{equation}
    \mathbf{h}_{1:T'} = \text{speech\_encoder}(\mathbf{X}_{1:T})
\end{equation}

We add an output layer $W_o$ to the speech encoder and fine-tune the model with CTC loss \cite{graves2006connectionist} on supervised ASR training data. The model output then becomes $\mathbf{o}_{1:T'}$,
\begin{equation}
\begin{split}
    \mathbf{z}_{t} &= W_o \cdot \mathbf{h}_{t} \\
    \mathbf{o}_{t} &= \text{softmax}(\mathbf{z}_{t})
\end{split}    
\end{equation}
The CTC loss function is given by:
\begin{equation}
    \mathcal{L}_{\text{CTC}}(\mathbf{y}_{1:N},\mathbf{o}_{1:T'})=-\log \sum_{\pi\in\mathcal{A}(\mathbf{y})} \prod_{t=1}^{T'}\mathbf{o}_{t}^{(\pi_t)}
\end{equation}
where $\mathcal{A}(\mathbf{y})$ represents the set of all valid alignments of generating the target sequence $\mathbf{y}$ from the input sequence $\mathbf{X}$.

For the CTC training process, we use the same vocabulary as the one used by LLM. Let $|V|$ be the size of the LLM vocabulary, the CTC outputs thus consist of $|V|+1$ tokens, including the ``blank'' token. In the experimental section, we show that our approach remains effective when the vocabularies of the speech encoder and the LLM differ. 

\subsection{The LegoSLM Connection Method}

\begin{figure}[!htbp]
    \centering
    \includegraphics[width=1.0\linewidth]{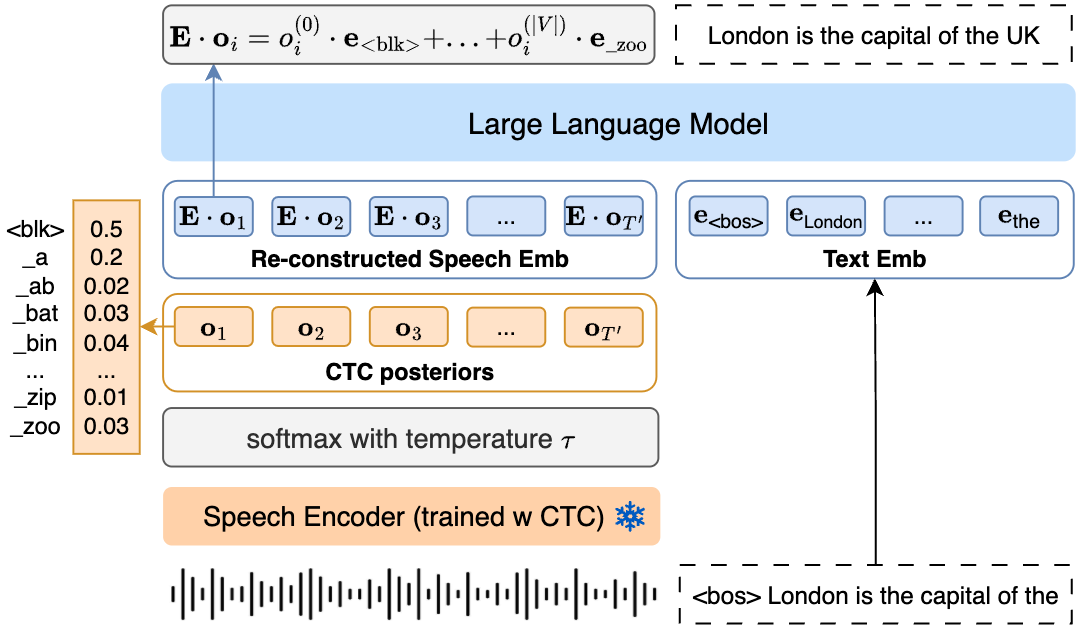}
    \caption{Depiction of the proposed LegoSLM method. The speech embeddings are reconstructed using ASR CTC posteriors and the LLM embedding table.}
    \label{fig:legoslm}
\end{figure}

The architecture of the LegoSLM system is illustrated in Figure \ref{fig:legoslm}, where the generated CTC posteriors $\mathbf{o}_{1:T'} $ are utilized to integrate the speech encoder with the LLM.
In the adaption, we freeze the model weights of the speech encoder and only fine-tune the LLM. 
To generate speech representations $\mathbf{s}_{t}$ that are aligned with the text embedding space, we compute a weighted sum of the LLM embedding table $\mathbf{E}$ from the CTC posteriors $\mathbf{o}_{1:T'}$,
\begin{equation}
    \mathbf{s}_{t} = \mathbf{E} \cdot \mathbf{o}_{t} \\
\end{equation}
These computed speech embeddings are concatenated with the text embeddings for processing in subsequent layers. 

Compared to the speech prompt method, we also use continuous vectors to represent information from the original utterance. However, the text embeddings are used as codebooks for speech embedding reconstruction, which implicitly matches the two modalities.
In this paper, we showcase the application of LegoSLM architecture in ASR and AST tasks. For ASR, LLM is trained to generate transcriptions based on the CTC posterior outputs from the speech encoder. For AST, the speech encoder is trained to generate posteriors over tokens of the source language, which are used by the LLM to produce translations in the target language.

\subsection{Discussion of the CTC Blank Token}
CTC network outputs a special token \texttt{<blk>} for use in the alignment process, which is not part of the LLM vocabulary. In the experiments, we map it to a new LLM embedding vector $\mathbf{e}_{\text{<blk>}}$ that is randomly initialized. 
Usually, the blank tokens occur more frequently than the non-blank symbols in the CTC decoding result and tend to have sharper posteriors \cite{graves2006connectionist}.
As the \texttt{<blk>} token carries limited content information about the utterance, we introduce a model variant that suppresses its probability in the $\mathbf{z}_t$ distribution, thereby enhancing the representation of more meaningful tokens.
\begin{equation}
\begin{split}
    \mathbf{\hat{z}}_t^{\text{<blk>}}=\mathbf{z}_t^{\text{<blk>}}-\log(\text{blk\_downscale})
\end{split}
\end{equation}
The default value of \texttt{blk\_downscale} is set to 1 in the experiments. In a model variant LegoSLM*, we increase \texttt{blk\_downscale} to $1e4$, and further increasing it does not lead to improved performance.








\subsection{Zero-shot System Combination}

\begin{figure}[!htbp]
    \centering
    \includegraphics[width=0.86\linewidth]{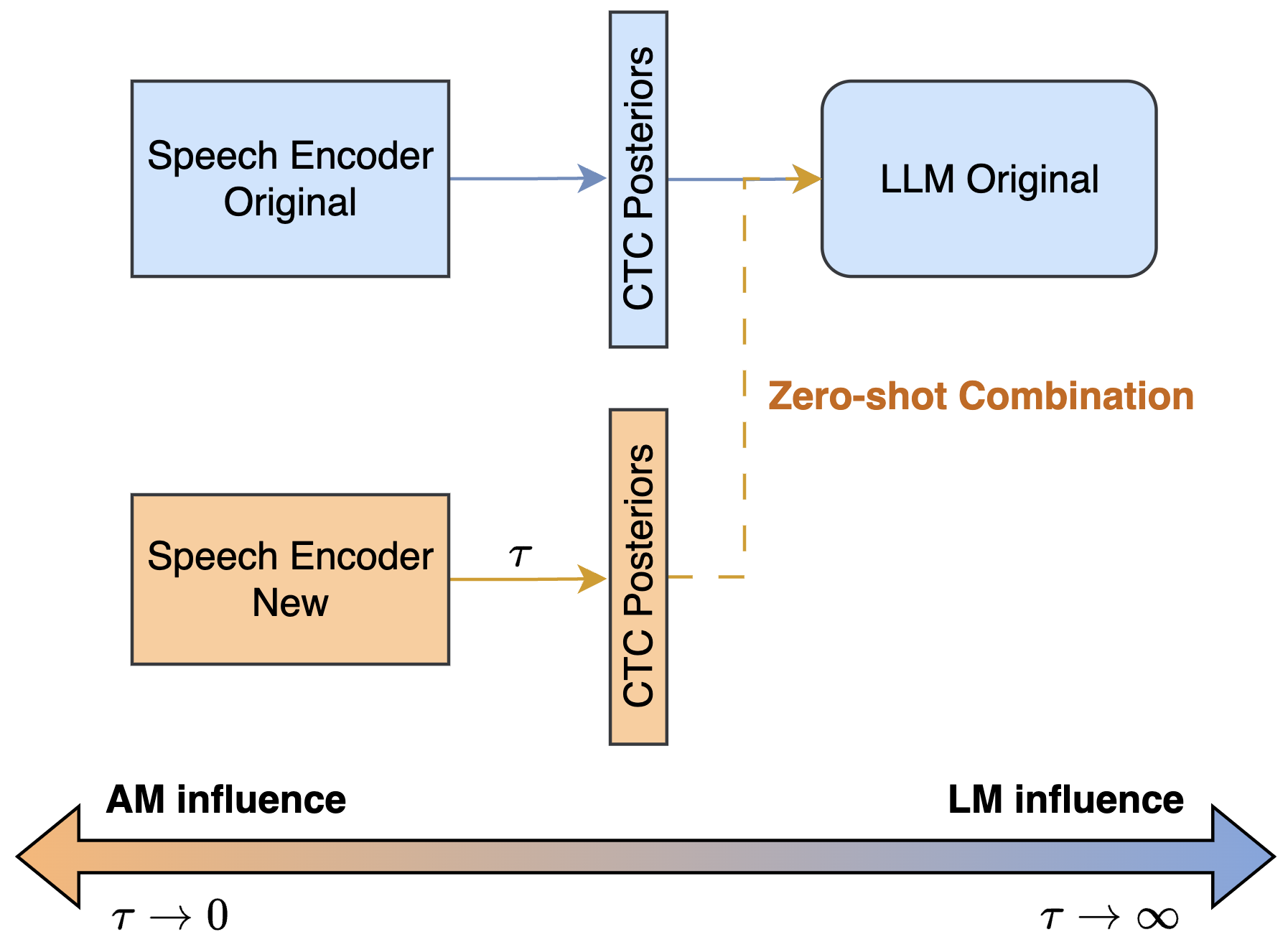}
    \caption{Illustration of the zero-shot system combination test where speech encoders and LLMs trained in different setups are seamlessly combined. 
    }
    \label{fig:combine}
\end{figure}

An advantage of the LegoSLM architecture is that it enforces the modularity between the speech encoder and the LLM since only CTC posteriors are passed to the LLM. 
After training the system, the LLM can accept outputs from a different speech encoder as long as the model also operates on the same CTC vocabulary.
This property is useful in real-life applications, where, for example, the ASR encoder needs to be updated with more training data or adapted to a new domain. In the experiments, we conduct the zero-shot system combination test, as illustrated in Figure \ref{fig:combine}. Specifically, we fine-tune the LLM weights based on outputs from the original speech encoder. Then during evaluation, we plug in a separately-trained speech encoder to the LLM decoder and evaluate the system performance without updating any model weight.

\subsection{AM/LM Spectrum Control}

Following the previous section, when we seamlessly combine speech encoder and LLM trained in different setups, the capability of these models may differ. In traditional phonetic-based systems, the acoustic model (AM) and the language model (LM) are trained individually and combined during decoding. 
To achieve better recognition performance, a scaling parameter is often introduced to adjust the influence of the two models in the system combination \cite{young2002htk}. 
A similar strategy is employed for integrating external language models into E2E ASR systems \cite{toshniwal2018comparison}. 
Nevertheless, for work that empowers LLM with speech ability, how to control the weighting of the speech encoder and the LLM has not been thoroughly explored in previous studies.
To address this research question, we propose to use a temperature parameter $\tau > 0$ in the CTC softmax layer for influence control,
\begin{equation}
    o_t^{(i)} = \frac{\exp(z_t^{(i)} / \tau)}{\sum_{j=1}^{|V|+1} \exp({z_t^{(j)} / \tau})}
\end{equation} 
When $\tau>1$, the CTC probabilities become flatter, allowing greater flexibility for the LLM in the generation process. Conversely, when $\tau<1$, the CTC probability distribution becomes sharper, granting the speech encoder more certainty.


\subsection{Model Variants}
In the LegoSLM method, the CTC probability distribution over the entire vocabulary is utilized. 
Nonetheless, most probability is distributed to the top token predictions, following a long-tail distribution. Therefore, retaining only the top $K$ predictions at each frame's output is expected to preserve most of the information.
In the following, two model variants are introduced: combining the top-$K$ CTC predictions with softmax (LegoTopS) and with a projection layer (LegoTopP).

In the LegoTopS approach, a vector $\mathbf{i}_t$ is computed that corresponds to the indices of the top-$K$ values in the speech encoder output $\mathbf{z}_t$. Accordingly, we reconstruct the speech embeddings $\tilde{\mathbf{s}}_{1:T'}$ by computing a weighted sum of the associated LLM input embeddings using the top-$K$ predictions as multipliers,
\begin{equation}
    \begin{split}
    \mathbf{i}_t &= \text{argmax\_k} (\mathbf{z}_t) \\
    \tilde{\mathbf{s}}_t &= E[\mathbf{i}_t] \cdot \text{softmax}(\mathbf{z}_t[\mathbf{i}_t])
    \end{split}
\end{equation}

With the LegoTopP variant, LLM token embeddings associated with the $\mathbf{i}_t$ predictions are concatenated and mapped to the original embedding dimension using a linear projection layer. This layer is randomly initialized and jointly trained with the LLM weights in the adaptation. Different values of $K$ are tested in the experimental section.


\section{Experimental Setup}

\subsection{Models}

To demonstrate the effectiveness of our proposed LegoSLM framework, we design experiments on both ASR and AST tasks. The Universal Speech Model (USM) \cite{zhang2023google} with 300M parameters is used as the speech encoder in our experiments. The model is pre-trained on multilingual YouTube data with the BEST-RQ loss in an unsupervised fashion \cite{chiu2022self}. We build a USM-CTC model by adding a CTC layer to the USM encoder and fine-tune the model on various datasets to simulate scenarios with varying amounts of available labeled ASR data. For the LLM decoder, we experiment with the Gemma 2B model \cite{team2024gemma}. The \texttt{pt\_f32} and \texttt{it\_f32} checkpoints are used separately in the ASR and AST experiments. The former model is only trained with predicting the next token while the latter one is further trained with instruction tuning. In the LegoSLM training, we freeze the USM-CTC model and only adapt the LLM weights. 
The detailed training setup and hyperparamers can be found in Appendix \ref{app:training}.

\subsection{Baselines}
To evaluate the performance of the proposed method, several baseline approaches are compared:

\noindent\textbf{USM-CTC:} Since the USM is trained with CTC loss, we can decode it on the test set and compute the WER for the speech encoder solely. In the experiment section, we show the beam search results with a beam size of 10.

\noindent\textbf{Speech Prompts (SP):} For the speech prompt method, we test a simple yet effective mapping network to bridge the speech encoder and the LLM -- training a linear layer in between for dimension match \cite{fathullah2024prompting}. Our preliminary experiments suggest that Gemma tends to hallucinate when its weights are frozen, likely because the Gemma model was trained without any speech inputs. Hence, in the \texttt{U+P+G} setup, we update weights of USM, projection layer, and the Gemma model. We also train with the \texttt{P+G} setup where only the projection layer and the Gemma model are tuned. Moreover, as indicated by preliminary experiments, stacking multiple speech embeddings to reduce the input length does not yield performance improvements. Consequently, speech outputs from each frame are fed as individual LLM inputs.

\noindent\textbf{ASR Error Correction (AEC):} For AEC, the ASR N-best list generated by the beam search on USM-CTC is collected and fed as input to train the LLM decoder. Following \cite{ma2023nbest}, top $n$-best ASR hypotheses are concatenated in order as the model input, separated by \texttt{<sep>} tokens in between to denote the sentence boundaries.

\subsection{Dataset}

For ASR experiments, we train USM-CTC models and the corresponding Gemma models in four different settings: three English-only ASR systems and a multilingual system built for 8 languages. The \textbf{mls-en} model is trained on the en-us part of the MLS \cite{pratap20_interspeech} dataset. \textbf{Public} represents the combination of the MLS en-us subset and the SpeechStew dataset \cite{chan2021speechstew}, which is a collection of multiple public ASR datasets. The model with the \textbf{lbs} label is only trained on the LibriSpeech training set \cite{panayotov2015librispeech}. In the multilingual setup, model \textbf{multi} learns from MLS training data of all 8 languages. To evaluate the model performance, WER results on the English test set from MLS and the test\_other set from LibriSpeech are calculated, denoted as \texttt{MLS\_en} and \texttt{LBS\_other}. For speech translation, we conduct experiments on the public CoVoST 2 dataset \cite{wang2021covost}. We use the same USM-CTC models in the \texttt{multi} ASR experiments and train Gemma models separately on three translation directions: \texttt{fr$\rightarrow$en}, \texttt{de$\rightarrow$en} and \texttt{en$\rightarrow$de}. All datasets are publicly available for research purposes, and their use in this paper aligns with their intended purpose.

\section{Results}

\subsection{Experiments on ASR}


Table \ref{tab:mls-en} listed the ASR performance for models trained on MLS en-us data. 
For the speech prompt method, the best performance can be observed on both sets when all components including the USM model weights, projection layer, and the Gemma weights are jointly fine-tuned. Since the USM weights are kept frozen in the LegoSLM adaptation phase, our proposed method is more comparable to the \texttt{P+G} setup. 
For AEC, $n=1$ leads to minor performance improvement on the MLS\_en set and degradation on the LBS\_other set. The results indicate that relying solely on the top-1 ASR hypothesis limits the LLM's ability to correct errors effectively. To achieve better performance, we feed ASR N-best lists as input, which include highly probable transcription alternatives. Increasing $n$ leads to better performance, and when the 10-best list is used, the best performance of 13\% WERR can be seen on the MLS\_en set.

\begin{table}[!htbp]
    \centering
    \small
    \begin{tabular}{@{ }p{1.5mm}|l|c|c}
    \toprule
        \multicolumn{2}{c|}{Model (mls-en)} & MLS\_en & LBS\_other \\
    \midrule
        \multicolumn{2}{l|}{Encoder: USM-CTC} & 8.9 & 6.8 \\
    \midrule
        \multirow{8}*{\rotatebox{90}{+Gemma}} & SP \texttt{(U+P+G)} & 5.2 & 4.8 \\
        & SP \texttt{(P+G)} & 5.5 & 5.2 \\
        \cmidrule{2-4}
        & AEC \texttt{(n=1)} & 8.9 & 8.0 \\
        & AEC \texttt{(n=5)} & 8.0 & 6.5 \\
        & AEC \texttt{(n=10)} & 7.8 & 5.7 \\
        \cmidrule{2-4}
        & Ours: LegoSLM & 6.1 & 5.5 \\
        & Ours: LegoSLM* & \textbf{5.6} & \textbf{5.2} \\
    \bottomrule
    \end{tabular}
    \caption{WER results for speech prompts, ASR error correction, and the proposed LegoSLM method. Models are trained on the MLS en-us data. Results with * reduce the predicted probability of the \texttt{<blk>} token.}
    \label{tab:mls-en}
\end{table}

\begin{table*}[!htbp]
    \centering
    \small
    \begin{tabular}{@{ }p{1.5mm}|l|cccccccc|c}
    \toprule
        \multicolumn{2}{c|}{Model (multi)} & en & de & nl & fr & es & it & pt & pl & \textbf{Avg.} \\
    \midrule
        \multicolumn{2}{l|}{Baseline: USM-CTC} & 9.5 & 12.7 & 18.2 & 14.5 & 10.2 & 23.0 & 22.5 & 32.1 & 17.8 \\
    \midrule
        \multirow{3}*{\rotatebox{90}{+Gemma}} & SP \texttt{(P+G)} & 5.5 & 6.1 & 12.2 & 5.4 & 4.9 & 11.6 & 12.7 & 11.7 & 8.8 \\
        & AEC \texttt{(n=10)} & 8.2 & 9.5 & 14.5 & 9.2 & 8.0 & 19.1 & 19.1 & 22.9 & 13.8 \\
        \cmidrule{2-11}
        & Ours: LegoSLM* & 5.7 & 5.1 & 12.1 & 5.8 & 5.1 & 12.4 & 13.6 & 12.7 & \textbf{9.1} \\
    \bottomrule
    \end{tabular}
    \caption{WER results for models trained on the multilingual MLS data across 8 languages.}
    \vspace{-2mm}
    \label{tab:multiple}
\end{table*}

Our proposed LegoSLM method achieves a WER of 6.1 on the MLS\_en test set. Here, we use the probability predicted in the CTC layer output directly.
With the LegoSLM* setting, the logit of the \texttt{<blk>} token is decreased by $\log(1e4)$ to reduce its influence.
Results indicate that decreasing the weight of \texttt{<blk>} when reconstructing the speech embeddings leads to notable performance improvement, contributing to a total WERR of 37\%. 
The LegoSLM models largely outperform AEC and are more cost-effective when generating features. In the AEC approach, the input length increases proportionally with $n$, leading to training inefficiencies. Additionally, generating the ASR N-best list involves running a beam search, whereas LegoSLM only uses the plain probability distribution. LegoSLM* also achieves comparable performance with the SP \texttt{(P+G)} setting. These findings suggest that using CTC posteriors as an intermediate representation preserves most of the information compared to the continuous ASR outputs. The WER decomposition is listed in Table \ref{tab:wer_decom}.

Table \ref{tab:multiple} presents ASR performance on the multilingual LibriSpeech dataset, where the model is jointly trained on speech data from eight languages. The projected speech prompts, ASR error correction, and LegoSLM* achieve average WERRs of 50\%, 22\%, and 49\%, respectively. As indicated by the results, our proposed method demonstrates strong performance in the multilingual setup. Nevertheless, the performance gap of LegoSLM* with the SP method widens for languages with less training data, such as Polish with 104h of speech data. Additional results on LibriSpeech and public English datasets, provided in Appendix \ref{other_sets}, further validate the effectiveness of our approach.


\subsection{Zero-shot System Combination}

Table \ref{tab:zeroshot} evaluates the modularity of various connection methods. In this setup, the Gemma model is initially trained using outputs from the USM-CTC encoder that is developed on the MLS en-us dataset. This is the same model setting in Table \ref{tab:mls-en}. During the evaluation, this trained Gemma model is utilized without additional training to integrate outputs from other speech encoders. 
As shown in the table, the continuous speech prompt method fails to transfer across setups, as the output spaces from different speech encoders are incompatible, causing the LLM to underperform. For the cascaded AEC method, the trained Gemma model shows effectiveness in refining the transcriptions from other speech encoders, though the performance gains over the USM-CTC baseline remain limited.

\begin{table}[!htbp]
    \centering
    \small
    \begin{tabular}{@{ }p{1.5mm}|l|c|c|c}
    \toprule
        \multicolumn{2}{c|}{Model} & multi & public & lbs \\
    \midrule
        \multicolumn{2}{l|}{Encoder: USM-CTC} & 9.5 & 10.7 & 12.8 \\
    \midrule
        \multirow{8}*{\rotatebox{90}{+Gemma (\text{mls-en})}} & SP \texttt{(U+P+G)} & 200.0 & 165.2 & 218.6 \\
        & SP \texttt{(P+G)} & 195.9 & 169.3 & 181.8 \\
        \cmidrule{2-5}
        & AEC \texttt{(n=1)} & 9.6 & 10.0 & 12.2 \\
        & AEC \texttt{(n=5)} & 8.8 & 9.2 & 11.0 \\
        & AEC \texttt{(n=10)} & 8.3 & 8.6 & 10.7 \\
        \cmidrule{2-5}
        & Ours: LegoSLM & 6.7 & 7.2 & \textbf{8.1} \\
        & Ours: LegoSLM* & \textbf{6.4} & \textbf{7.0} & 8.8 \\
    \bottomrule
    \end{tabular}
    \caption{Experimental results of zero-shot system combination on the MLS\_en test set. The Gemma model is trained in the mls-en setup while we plug in USM encoders trained from multi, public, and lbs setups.}
    \label{tab:zeroshot}
\end{table}

The LegoSLM models maintain strong performance across different USM encoders, achieving WERRs of 32\% to 37\% compared to baseline CTC decoding results. For the USM-CTC model trained on LibriSpeech, reducing the weight of the \texttt{<blk>} token leads to performance degradation, likely due to the distributional differences between LibriSpeech and the MLS en-us dataset. As a result, using the original probability distribution eases the transfer process. 
These experiments highlight LegoSLM's modularity, a valuable property in real-life applications where the LLM does not require retraining when the ASR encoder is replaced.



\subsection{Experiments on AM/LM Spectrum}
\label{sec:zero-shot}


\begin{figure}[!htbp]
    \centering
    \includegraphics[width=0.99\linewidth]{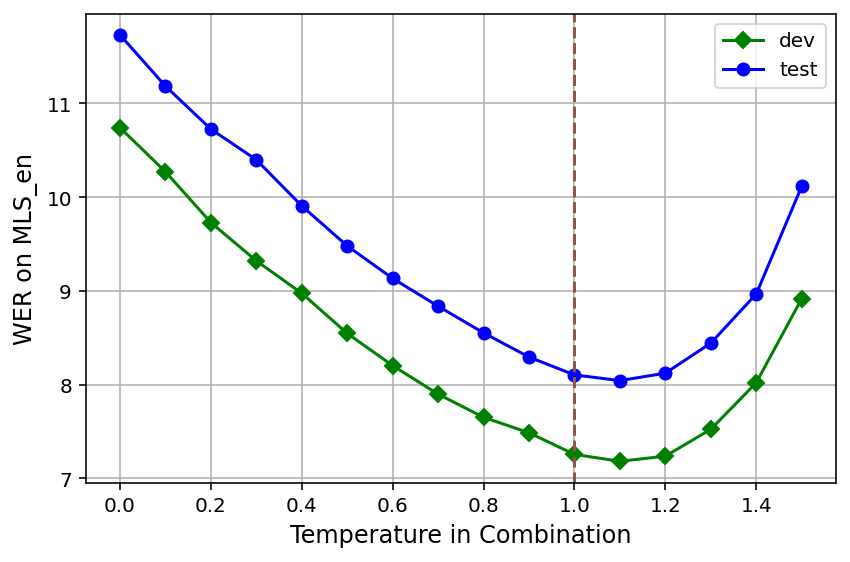}
    \includegraphics[width=0.99\linewidth]{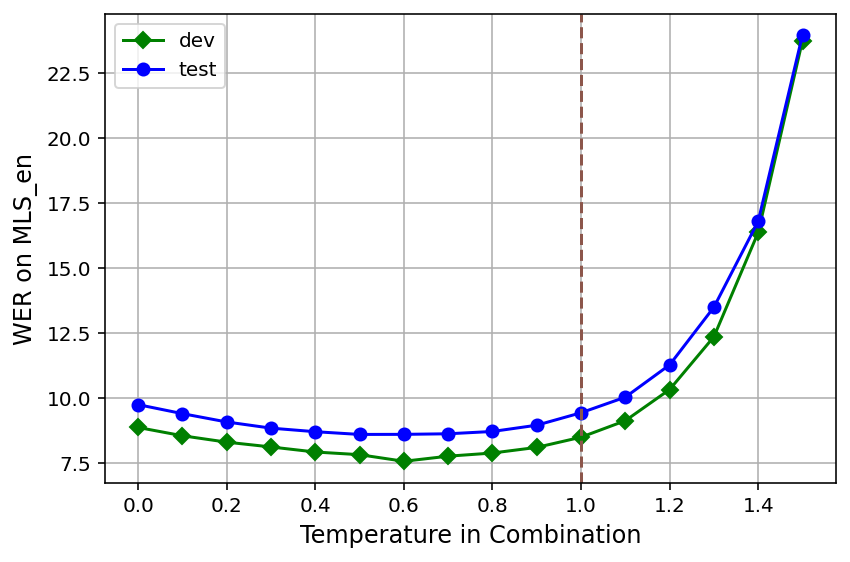}
    \caption{Effect of changing the temperature value in LegoSLM. The speech encoder and LLM are trained on different setups. Top: USM-CTC (lbs) + Gemma (mls-en). Bottom: USM-CTC (mls-en) + Gemma (lbs). }
    \label{fig:temp}
\end{figure}
In the context of zero-shot system combination, 
the encoder and decoder may exhibit varying levels of capability when trained on different data. 
In the following experiments, we utilize the temperature value $\tau$ in the CTC softmax to adjust the emphasis given to the speech encoder and LLM in the generation.
Figure \ref{fig:temp} demonstrates the impact of selecting different temperature values during softmax computation. 
In the first setup, the USM model was trained on LibriSpeech, while the Gemma decoder was fine-tuned on MLS-en. Since the decoder is more robust in this scenario, the model shows the best performance at $\tau=1.1$, granting the LLM greater freedom during decoding.
In contrast, for the setup depicted in the figure below, the optimal temperature value is approximately 0.6, emphasizing the role of the USM-CTC encoder in the system combination. These results highlight the effectiveness of temperature control in balancing the contributions of the encoder and decoder, leading to improved overall ASR performance.
Detailed WER results and a case study 
illustrating the effects of temperature values 
can be found in Appendix \ref{app:temp}.




\subsection{Model Variants}

\begin{table}[!htbp]
    \centering
    \small
    \begin{tabular}{l|c|cc}
    \toprule
    Model & $K$ & MLS\_en & LBS\_other \\
    \midrule
    LegoSLM* & - & \textbf{5.6} & \textbf{5.2} \\
    \midrule
    \multirow{3}*{LegoTopP*} & 3 & 6.2 & 5.9 \\
    & 5 & 6.1 & 6.7 \\
    & 10 & 6.0 & 7.5 \\
    \midrule
    \multirow{3}*{LegoTopS*} & 1 & 7.0 & 6.6 \\
    & 10 & 5.8 & 5.5 \\
    & 100 & \textbf{5.6} & 5.3 \\
    \bottomrule
    \end{tabular}
    \caption{Ablation of the LegoSLM* approach where we only keep the top-$K$ token predictions at each frame.}
    \label{tab:top_n}
\end{table}

In Table \ref{tab:top_n}, we analyze the impact of constraining the LLM input to be the top-$K$ tokens derived from the CTC posteriors at each frame. Results at $K=10$ indicate that using a softmax-based approach to combine LLM text embeddings outperforms the projection-based method, emphasizing the importance of incorporating ASR scores. At $K=100$, the performance of LegoTopS* is comparable to LegoSLM*, indicating robustness across varying input sizes. Notably, due to the long-tail distribution of the CTC outputs, retaining only the top token predictions preserves the most relevant information, leading to minimal performance degradation compared to using the probability distribution generated on the full vocabulary.
This method also has a speed advantage, as softmax over the entire vocabulary is no longer required.
In the extreme case of $K=1$, which corresponds to utilizing CTC greedy decoding results without token merging or blank removal, the ASR performance drops substantially. This highlights the importance of leveraging alternative predictions for the LLM to generate accurate ASR hypotheses.




\begin{table}[!htbp]
    \centering
    \small
    \renewcommand\tabcolsep{5pt}
    \begin{tabular}{l|l|cc}
    \toprule
       CTC Vocab & Model & MLS\_en & LBS\_other \\
    \midrule
        256K & USM-CTC & 8.9 & 6.8 \\
        (matched) & Ours: LegoSLM* & \textbf{5.6} & \textbf{5.2} \\
    \midrule
        \multirow{1}*{16,384} & USM-CTC & 10.0 & 7.5 \\
        (different) & Ours: LegoSLM* & \textbf{5.6} & \textbf{5.2} \\
    \bottomrule
    \end{tabular}
    \caption{WER results for LegoSLM* systems trained on MLS en-us data. The Gemma model uses a vocabulary of 256K tokens and we test both cases when USM-CTC uses a matched and a different vocabulary.}
    \label{tab:diff}
\end{table}

Previous experiments were conducted under the assumption that the speech encoder and the LLM share the same vocabulary. However, this constraint may be difficult to meet in practical applications. 
Table \ref{tab:diff} further evaluates the system's performance when the vocabularies remain different. Specifically, the USM-CTC model is trained with a vocabulary size of 16K, while the Gemma model operates with 256K tokens.
To handle this mismatch, an additional input embedding table is introduced to the LLM. This table is randomly initialized and jointly trained alongside the other LLM parameters. These 12M additional parameters effectively map the ASR output logits to the LLM’s input space. Baseline results from CTC decoding indicate that the USM-CTC system performs better when trained with a fine-grained vocabulary. However, after fine-tuning using the LegoSLM* method, ASR performance becomes comparable, highlighting the robustness of our approach even when the ASR and LLM vocabularies are not aligned.


\subsection{Experiments on Speech Translation}

Table \ref{tab:ast} presents the experimental results of the speech translation task, which requires the model to comprehend the semantics of the utterance in the source language and generate accurate transcriptions in the target language. In the Oracle setup, a text translation system is trained using ASR references as input, serving as an upper bound for performance evaluation. For the other approaches, the USM-CTC model learns from the multilingual LibriSpeech data to perform speech recognition in the source language. The Gemma decoder is tuned on CoVoST 2 to generate translation in the target language given outputs from the speech encoder. When the training data is limited, aligning the speech encoder outputs with the LLM text embedding space becomes challenging, leading to the underperformance of the speech prompts method.
Nevertheless, the cascaded AEC systems exhibit strong performance, as utilizing the transcription in the source language assists the LLM in understanding the utterance's meaning, thus reducing the complexity of the task.
The LegoSLM* method further boosts system performance by mitigating information loss compared to AEC, achieving the best results across all AST configurations.

\begin{table}[h!]
    \centering
    \small
    \begin{tabular}{l|c|c|c}
    \toprule
         Model & fr$\rightarrow$en & de$\rightarrow$en & en$\rightarrow$de \\
         \midrule
         Oracle & 36.5 & 30.1 & 31.8 \\
    \midrule
        SP \texttt{(U+P+G)} & 11.3 & 9.8 & 15.7 \\
        SP \texttt{(P+G)} & 9.7 & 7.5 & 13.4 \\
        \midrule
        AEC \texttt{(n=1)} & 18.7 & 16.7 & 18.6 \\
        AEC \texttt{(n=5)} & 20.7 & 18.2 & 19.8 \\
        AEC \texttt{(n=10)} & 21.5 & 18.7 & 20.3 \\
        \midrule
        Ours: LegoSLM & 23.8 & 18.9 & 20.3 \\
        Ours: LegoSLM* & \textbf{25.8} & \textbf{21.1} & \textbf{21.1} \\
    \bottomrule
    \end{tabular}
    \caption{BLEU scores ($\uparrow$) for speech translation performance on CoVoST 2 test sets. Oracle fine-tunes the Gemma model using ASR reference texts as input.}
    \label{tab:ast}
\end{table}

\begin{table}[!htbp]
    \centering
    \small
    \begin{tabular}{l|c|c|c}
    \toprule
        Model & mls-en & public & lbs \\
    \midrule
        SP \texttt{(U+P+G)} & 3.5 & 3.0 & 2.8 \\
        SP \texttt{(P+G)} & 3.1 & 2.9 & 4.0 \\
        \midrule
        AEC \texttt{(n=1)} & 19.5 & 21.4 & 17.7 \\
        AEC \texttt{(n=5)} & 20.6 & 22.2 & 18.3 \\
        AEC \texttt{(n=10)} & \textbf{20.9} & 22.3 & 18.4 \\
        \midrule
        Ours: LegoSLM (best) & \textbf{20.9} & \textbf{22.4} & \textbf{18.7} \\
    \bottomrule
    \end{tabular}
    \caption{BLEU scores of the zero-shot system combination on the \texttt{en$\rightarrow$de} AST test set. 
    }
    \label{tab:ast_modular}
\end{table}




For the AST task of translating English utterances into German text, we test the modularity of the system by swapping the USM encoder. As shown in Table \ref{tab:ast_modular}, after training the Gemma model on the AST data, it becomes possible to seamlessly integrate a different USM-CTC encoder trained on another ASR set. For the LegoSLM method, we report the best BLEU score achieved with the optimal temperature value. This method demonstrates superior performance across three setups, with the AEC system using a 10-best list delivering comparable speech translation performance.

\begin{table}[!htbp]
    \centering
    \scriptsize
    \renewcommand\tabcolsep{5pt}
    \begin{tabularx}{\linewidth}{l|X}
    \toprule
    Type & Text \\
    \midrule
    ASR-REF & La France est un grand pays, nous en sommes convaincus. \\
    \midrule
    AST-REF & France is a big country, we are convinced about it. \\
    \midrule
    SP \texttt{(U+P+G)} & France is a conquered country, we are getting rid of it. \\
    \midrule
    LegoSLM* & France is a great country, we are convinced of it. \\
    \midrule\midrule
    ASR-REF & Nous avons décidé, au contraire, de renforcer la progressivité de l’impôt sur le revenu. \\
    \midrule
    AST-REF & We have decided, on the contrary, to strength the progressiveness of the income tax. \\
    \midrule
    SP \texttt{(U+P+G)} & We had decided to give flexibility to companies. \\
    \midrule
    LegoSLM* & On the contrary, we have decided to strength the income tax. \\
    \bottomrule
    \end{tabularx}
    \caption{Case analysis on the \texttt{fr$\rightarrow$en} AST test set.}
    \label{tab:ast_case}
\end{table}

Table \ref{tab:ast_case} presents two examples from the AST test set. The speech translation results produced using the continuous speech prompt approach fail to accurately convey the sentence meaning, whereas the proposed LegoSLM* method generates translations that closely align with the reference text.

\section{Conclusions}
In this work, we propose a novel approach to combine a pre-trained speech encoder and LLM.
Extensive experimental results show that LegoSLM achieves competitive performance compared to prior approaches.
Since CTC posteriors are used to bridge the two modules, after training the LLM, the ASR encoder can be switched in a zero-shot manner. 
Additionally, we propose using the temperature value in softmax to adjust the relative emphasis placed on the speech encoder and the LLM components. Furthermore, several model variants are introduced and evaluated.
The results presented in this paper indicate that LegoSLM has potential for broader applications, such as speech summarization and spoken language understanding.



\section{Limitations}

This study serves as an initial exploration of how the CTC posteriors from a speech encoder empower LLMs to handle the speech modality. In this paper, we present experiments using the USM and Gemma models to demonstrate the effectiveness of our approach. 
We generate posteriors from a CTC-based speech encoder, given its efficiency and its popularity in the speech pre-training field. Nevertheless, our approach can be extended to speech encoders with other architectures such as RNN-T or LAS models. 
Moving forward, we aim to expand our analysis to other large-scale foundation models to draw broader conclusions. In this work, we employ fine-tuning to adapt the LLM weights. However, alternative parameter-efficient tuning methods, such as LoRA, are commonly used for adapting LLMs. While this is not addressed in the current version, we anticipate observing similar performance trends with these methods.

\section{Risks and Ethics}
There are no known ethical concerns or risks associated with the findings of this work.

\bibliography{custom}

\clearpage
\appendix

\section{Training Details}
\subsection{Dataset Details}
\label{app:data}

For speech recognition, MLS \cite{pratap20_interspeech}, LibriSpeech \cite{panayotov2015librispeech} and SpeechStew \cite{chan2021speechstew} datasets are used in our experiments. The LibriSpeech corpus consists of 960 hours of English read speech data, collected from the LibriVox project. Later on, MLS is released, which is a multilingual version of LibriSpeech at a larger scale. It is also derived from audiobooks and contains data from 8 languages, with 44.5K hours of English data and 6K hours of speech over other 7 languages. Speechstew is a large-scale multi-domain ASR dataset created by mixing various public datasets: AMI, Broadcast News, Common Voice, LibriSpeech, Swithboard/Fisher, and WSJ.

The standard CoVoST 2 \cite{wang2021covost} corpus is used for training and evaluation in the speech translation experiments. It is a diversified translation set based on the Common Voice project \cite{ardila2020common}. For each language pair, the speech from the source language and the text reference in the target language is provided. The original data contains translation from 21 languages into English and translation from English into 15 languages. In this paper, we design experiments in three translation directions. The number of utterances and hours of speech data are listed in Table \ref{tab:data_details}.

\begin{table}[!htbp]
    \centering
    \small
    \begin{tabular}{l|l|c|c|c}
    \toprule
        Task & Data & Split & \#Utts & Hours \\
    \midrule
        \multirow{10.5}*{ASR} 
        & \multirow{8}*{MLS} & en-us & 10,808K & 44,660 \\
        & & de-de & 469K & 1,966 \\
        & & nl-nl & 374K & 1,554 \\
        & & fr-fr & 258K & 1,076 \\
        & & es-es & 220K & 917 \\
        & & it-it & 59K & 247 \\
        & & pt-br & 37K & 161 \\
        & & pl-pl & 25K & 104 \\
    \cmidrule{2-5}
        & LibriSpeech & - & 281K & 960 \\
    \cmidrule{2-5}
        & SpeechStew & - & 4,452K & 4,730 \\
        \midrule
        \multirow{3}*{AST} & \multirow{3}*{CoVoST 2} & fr$\rightarrow$en & 207K & 180 \\
        & & de$\rightarrow$en & 127K & 119 \\
        & & en$\rightarrow$de & 289K & 364 \\
    \bottomrule
    \end{tabular}
    \caption{Statistics of the training datasets.}
    \label{tab:data_details}
\end{table}
\vspace{-2mm}
\subsection{Training Details}
\label{app:training}

The USM-CTC models are fine-tuned from the pre-trained USM BERT-RQ checkpoints on multiple ASR training sets. The encoder contains 2 layers of subsampling convolution layers and 24 Conformer layers, with a model dimension of 768. In the fine-tuning, a linear layer with softmax is added to the USM encoder. The model is trained with CTC loss to make predictions for each frame in a vocabulary of 256K. During the training, the learning rate increases linearly to the maximum of 3e-5 in 5K steps and decays exponentially to 5e-5. USM-CTC is trained for 200K steps with a batch size of 192 on the training set.
For the LLM used in our experiments, Gemma 2B has 18 Transformer layers with a model dimension of 2048. The vocabulary consists of 256K tokens, which is the same one used in the USM-CTC training. In the Gemma fine-tuning, a learning rate of 1e-4 with the cosine decay strategy is applied. Models are trained for 5000 steps on LibriSpeech and CoVoST 2 with a batch size of 512. The \textit{mls-en}, \textit{multi}, and \textit{public} models are trained for 25K, 40K, and 40K steps accordingly, with a batch size of 1024. 
For better generalization, SpecAugment \cite{park19e_interspeech} and a dropout rate of 0.1 are applied in training. 
All models are trained and tested on TPU pods.

For ASR training, no prompt is applied in the input prefix. In the AST training, several prompts are used interchangeably: 
1. \textit{Translate the \{src\} speech into \{tgt\} text:} 2. 
\textit{Translate this \{src\} audio into \{tgt\} text:} 3. \textit{Convert this \{src\} audio recording into \{tgt\} text:}
4. \textit{Transform this \{src\} audio into \{tgt\} text:}
5. \textit{Generate a \{tgt\} text version of this \{src\} audio:}
6. \textit{Extract the spoken content from this \{src\} audio and present it in \{tgt\} text:}
7. \textit{Turn this {\{src\}} audio into {\{tgt\}} text:}
8. \textit{Rewrite this {\{src\}} audio in {\{tgt\}} text:}
9. \textit{I need this {\{src\}} audio translated into {\{tgt\}} text:}
10. \textit{Can you translate this {\{src\}} audio recording into {\{tgt\}} text?}
11. \textit{What would this {\{src\}} audio sound like in {\{tgt\}}?}
12. \textit{Translate the audio into {\{tgt\}} text:}. 
In the evaluation, the first prompt is applied to all test utterances.



\section{ASR Performance}

\subsection{Additional Experiments on MLS en-us}

\begin{table}[!htbp]
    \centering
    \small
    \begin{tabular}{@{ }p{1.5mm}|l|c|ccc}
    \toprule
        \multicolumn{2}{c|}{Model (mls-en)} & WER & Sub & Del & Ins \\
    \midrule
        \multicolumn{2}{l|}{Encoder: USM-CTC} & 8.9 & 6.7 & 1.4 & 0.8 \\
    \midrule
        \multirow{8}*{\rotatebox{90}{+Gemma}} & SP \texttt{(U+P+G)} & 5.2 & 3.6 & 0.8 & 0.7 \\
        & SP \texttt{(P+G)} & 5.5 & 4.0 & 0.8 & 0.7 \\
        \cmidrule{2-6}
        & AEC \texttt{(n=1)} & 8.9 & 6.1 & 1.3 & 1.5 \\
        & AEC \texttt{(n=5)} & 8.0 & 5.5 & 1.2 & 1.3 \\
        & AEC \texttt{(n=10)} & 7.8 & 5.3 & 1.2 & 1.3 \\
        \cmidrule{2-6}
        & Ours: LegoSLM & 6.1 & 4.4 & 0.9 & 0.7 \\
        & Ours: LegoSLM* & \textbf{5.6} & 4.0 & 0.9 & 0.7 \\
    \bottomrule
    \end{tabular}
    \caption{WER decomposition of different ASR models trained on the MLS en-us data.}
    \label{tab:wer_decom}
\end{table}

The USM-CTC and Gemma models in this section are developed using the MLS en-us data. Table \ref{tab:wer_decom} shows the WER breakdown on the test set in terms of substitution, deletion and insertion errors. Compared to the USM-CTC baseline, our proposed LegoSLM reduces all types of recognition errors, in particular substitution errors.

\begin{table}[!htbp]
    \centering
    \small
    \renewcommand\tabcolsep{5pt}
    \begin{tabular}{l|cc|cc}
    \toprule
    \multirow{2}*{Model} & \multicolumn{2}{c|}{Log Perplexity} & \multicolumn{2}{c}{Token Accuracy} \\
    & Init & Final & Init & Final \\
    \midrule
    SP \texttt{(U+P+G)} & 116.2 & 0.1 & 0.0 & 97.8 \\
    SP \texttt{(P+G)} & 116.2 & 0.1 & 0.0 & 97.7 \\
    \midrule
    AEC \texttt{(n=1)} & 1.4 & 0.2 & 82.1 & 96.4 \\
    AEC \texttt{(n=5)} & 1.0 & 0.1 & 87.3 & 97.3 \\
    AEC \texttt{(n=10)} & 1.0 & 0.1 & 87.4 & 97.4 \\
    \midrule
    Ours: LegoSLM & 5.9 & 0.1 & 25.9 & 97.7 \\
    Ours: LegoSLM* & 4.8 & 0.1 & 25.2 & 97.7 \\
    \bottomrule
    \end{tabular}
    \caption{Average log perplexity ($\downarrow$) and average token accuracy ($\uparrow$) results on the dev set. ``Init'' and ``Final'' refer to the statistics gathered before the training starts and after the training process is completed.}
    \label{tab:ppl}
    \vspace{-1mm}
\end{table}

The teaching-forcing practice is adopted in training where outputs from the speech encoder as well as the correct sentence history are given to the LLM to predict the next token. 
Under this setup, we compute the log perplexity of generating the reference text and the accuracy of the LLM in predicting the next tokens on the dev set, as presented in Table \ref{tab:ppl}. 
Before training begins, AEC achieves the best performance on both tasks, as LLM can leverage information from ASR hypotheses during generation. SP performs poorly on both tasks since LLM struggles to interpret the continuous outputs from the speech encoder without training. The proposed LegoSLM methods show strong performance, suggesting that LLM can effectively extract information from the reconstructed speech embeddings. As a result, LegoSLM reduces the adaptation complexity compared to the speech prompt method. After the adaptation, all models achieve quite low perplexity and high accuracy on the dev set.

\begin{table}[!htbp]
    \small
    \centering
    \begin{tabular}{c|c|c|c}
    \toprule
    dropout & specaug & freeze\_emb & MLS\_en \\
    \midrule
    & &&  7.0 \\
       \midrule
      \cmark & & & 6.7 \\
       \midrule
     & \cmark & & 5.7 \\
       \midrule
      \cmark & \cmark & & \textbf{5.6} \\
    \midrule
    \cmark & \cmark & \cmark & 5.8 \\
    \bottomrule
    \end{tabular}
    \caption{WER results on the MLS en-us test set with different training setups for LegoSLM*.}
    \label{tab:reg}
\end{table}

In Table \ref{tab:reg}, we conduct an ablation study on various training configurations for LegoSLM* trained on the MLS en-us data. 
For the default experimental setting in this paper, we apply a dropout rate of 0.1 to the LLM and use SpecAugment where two frequency masks with 44 bins and two time masks with a ratio of 0.1 are set.
This setup achieves a WER of 5.6 on the test set. The ablation results reveal that omitting these techniques leads to a substantial drop in model performance. Additionally, we test a setup where the LLM embedding layer is frozen during adaptation. The model achieves comparable performance to fine-tuning all parameters, indicating that freezing the embedding table does not adversely affect the model's capability to integrate the speech modality.



\label{sec:appendix}


\subsection{Experiments on Other Datasets}
\label{other_sets}

Table \ref{tab:mls-en} and Table \ref{tab:multiple} train both English and multilingual ASR models on the MLS data. As listed below, we further present the WER results of models trained on a public English dataset and the LibriSpeech corpus
in Table \ref{tab:pub} and Table \ref{tab:lbs}. Here, 49K hours and 1K speech data are used in the training, respectively. The results consistently demonstrate that LegoSLM achieves strong performance regardless of the training data size. 

\begin{table}[!htbp]
    \centering
    \small
    \begin{tabular}{@{ }p{1.5mm}|l|c|c}
    \toprule
        \multicolumn{2}{c|}{Model (public)} & MLS\_en & LBS\_other \\
    \midrule
        \multicolumn{2}{l|}{Encoder: USM-CTC} & 10.7 & 7.2 \\
    \midrule
        \multirow{8}*{\rotatebox{90}{+Gemma}} & SP \texttt{(U+P+G)} & 5.1 & 3.4  \\
        & SP \texttt{(P+G)} & 5.6 & 4.4  \\
        \cmidrule{2-4}
        & AEC \texttt{(n=1)} & 9.6 & 7.5   \\
        & AEC \texttt{(n=5)} & 8.7 & 6.0  \\
        & AEC \texttt{(n=10)} & 8.3 & 5.6 \\
        \cmidrule{2-4}
        & Ours: LegoSLM & 5.9 & 4.8  \\
        & Ours: LegoSLM* & \textbf{5.7} & \textbf{4.6}  \\
    \bottomrule
    \end{tabular}
    \caption{WER results for models trained on public data consisting of SpeechStew and MLS en-us split.}
    \label{tab:pub}
    \vspace{-2mm}
\end{table}

\begin{table}[!htbp]
    \centering
    \small
    \begin{tabular}{@{ }p{1.5mm}|l|c|c}
    \toprule
        \multicolumn{2}{c|}{Model (lbs)} & MLS\_en & LBS\_other \\
    \midrule
        \multicolumn{2}{l|}{Encoder: USM-CTC} & 12.8 & 8.2 \\
    \midrule
        \multirow{8}*{\rotatebox{90}{+Gemma}} & SP \texttt{(U+P+G)} & 8.5 & 5.6  \\
        & SP \texttt{(P+G)} & 10.1 & 6.8  \\
        \cmidrule{2-4}
        & AEC \texttt{(n=1)} & 16.6 & 10.8 \\
        & AEC \texttt{(n=5)} & 14.3 & 8.3 \\
        & AEC \texttt{(n=10)} & 13.7 & 7.7 \\
        \cmidrule{2-4}
        & Ours: LegoSLM & 11.0 & 7.2  \\
        & Ours: LegoSLM* & \textbf{10.5} & \textbf{7.0} \\
    \bottomrule
    \end{tabular}
    \caption{WER results for models trained on LibriSpeech.}
    \label{tab:lbs}
\end{table}

\onecolumn
\clearpage
\section{Full Results of AM/LM Spectrum Control}
\label{app:temp}
\begin{table*}[!htbp]
    \centering
    \small
    \renewcommand\tabcolsep{5pt}
    \begin{tabular}{l|p{1.2cm}c|c|c|c|c|c|c|c|c|c|c|c|c|c}
    \toprule
        \multirow{2}*{USM/Gemma} & \multirow{2}*{Dataset} & \multirow{2}*{Split} & \multicolumn{13}{c}{Temperature ($\tau$)} \\
        &&& 1e-4 & 0.5 & 0.6 & 0.7 & 0.8 & 0.9 & 1.0 & 1.1 & 1.2 & 1.3 & 1.4 & 1.5 & 1e4 \\
    \midrule
    \multirow{4}*{mls-en / lbs} & \multirow{2}*{MLS\_en} & dev & 8.9 & 7.8 & 7.8 & \textbf{7.8} & 7.9 & 8.1 & 8.5 & 9.1 & 10.3 & 12.3 & 16.4 & 23.7 & 318.2 \\
    && test & 9.4 & 8.6 & \textbf{8.6} & 8.6 & 8.7 & 8.9 & 9.4 & 10.0 & 11.3 & 13.5 & 16.8 & 23.9 & 319.5 \\
    \cmidrule{2-16}
    & \multirow{2}*{LBS\_other} & dev & 7.5 & 6.4 & 6.3 & \textbf{6.3} & 6.3 & 6.5 & 6.8 & 7.0 & 7.6 & 9.1 & 11.5 & 16.2 & 683.9 \\
    && test & 7.3 & 6.3 & \textbf{6.2} & 6.3 & 6.2 & 6.4 & 6.8 & 7.2 & 7.9 & 9.1 & 11.6 & 16.6 & 679.5 \\
    \midrule
    \multirow{4}*{lbs / mls-en} & \multirow{2}*{MLS\_en} & dev & 10.7 & 8.5 & 8.2 & 7.9 & 7.7 & 7.5 & 7.2 & \textbf{7.2} & 7.2 & 7.5 & 8.0 & 8.9 & 320.6 \\
    && test & 11.7 & 9.5 & 9.1 & 8.8 & 8.6 & 8.3 & 8.1 & \textbf{8.0} & 8.1 & 8.4 & 9.0 & 10.1 & 322.0 \\
    \cmidrule{2-16}
     & \multirow{2}*{LBS\_other} & dev & 8.3 & 6.8 & 6.7 & 6.7 & 6.3 & 5.9 & 6.0 & 6.0 & \textbf{5.8} & 6.1 & 6.7 & 7.4 & 234.8 \\
    && test & 8.8 & 7.0 & 6.7 & 6.7 & 6.5 & 6.5 & 6.4 & 6.4 & \textbf{6.3} & 6.7 & 7.2 & 8.0 & 240.6 \\


    \midrule
    \multirow{4}*{public / lbs} & \multirow{2}*{MLS\_en} & dev & 10.4 & 9.0 & 8.9 & \textbf{8.8} & 8.8 & 9.0 & 9.3 & 9.9 & 11.2 & 14.2 & 19.7 & 34.0 & 318.2 \\
    && test & 11.3 & 9.9 & 9.8 & \textbf{9.8} & 9.9 & 10.1 & 10.4 & 11.2 & 12.8 & 15.4 & 22.0 & 36.9 & 319.5 \\
    \cmidrule{2-16}
    & \multirow{2}*{LBS\_other} & dev & 7.9 & 6.6 & 6.7 & 6.6 & 6.7 & \textbf{6.6} & 6.8 & 7.1 & 7.7 & 9.0 & 11.5 & 16.8 & 683.3 \\
    && test & 8.0 & 6.8 & 6.8 & 6.7 & \textbf{6.7} & 6.7 & 6.9 & 7.3 & 7.8 & 9.0 & 11.0 & 15.4 & 679.1 \\
    \midrule
    \multirow{4}*{lbs / public} & \multirow{2}*{MLS\_en} & dev & 11.1 & 8.7 & 8.2 & 8.0 & 7.6 & 7.3 & 7.1 & \textbf{6.9} & 7.1 & 7.4 & 8.0 & 9.1 & 212.3 \\
    && test & 12.1 & 9.6 & 9.2 & 8.7 & 8.4 & 8.0 & 7.8 & \textbf{7.6} & 7.8 & 8.1 & 8.8 & 9.9 & 212.6  \\
    \cmidrule{2-16}
     & \multirow{2}*{LBS\_other} & dev & 8.3 & 6.3 & 6.1 & 5.8 & 5,7 & 5.6 & \textbf{5.4} & 5.4 & 5.5 & 5.6 & 6.3 & 7.1 & 130.0 \\
    && test & 8.3 & 6.5 & 6.2 & 6.0 & 5.7 & 5.6 & 5.5 & \textbf{5.5} & 5.7 & 5.9 & 6.5 & 7.4 & 131.4 \\
    
    \bottomrule
    \end{tabular}
    \caption{WER results of zero-shot system combination for LegoSLM with various temperature values applied in the CTC softmax layer. In each setup, the USM-CTC and Gemma models are trained from different data.}
    \label{tab:wer_combine}
\end{table*}

Table \ref{tab:wer_decom} presents the detailed WER results across four ASR setups when different temperature values $\tau$ are used in the generation. 
In a normal softmax distribution, the temperature value is set to 1.0, allowing equal influence from the speech encoder and LLM. As indicated by the WER results, manipulating the value of $\tau$ leads to improved model performance compared to using the default value of 1.0. When the speech encoder is trained on more data, the optimal $\tau$ falls below 1.0, generating a sharper probability distribution and thus representing more certainty in the speech embeddings. Conversely, when the LLM learns from more data, using a larger $\tau$ results in a more uniform distribution of the CTC posteriors, allowing more freedom to LLM in the generation. We also evaluate the extreme cases by setting $\tau$ to $1e{-4}$ and $1e4$. The first experiment closely resembles the USM-CTC baseline, while the second results in hallucination in the LLM generation. When $\tau$ is approaching infinity, speech embeddings become average vectors of the LLM embedding table and fail to convey meaningful information about the utterance.



\begin{figure}[!htbp]
  \centering
  \begin{minipage}{0.495\textwidth}
    \centering
    \includegraphics[width=\textwidth]{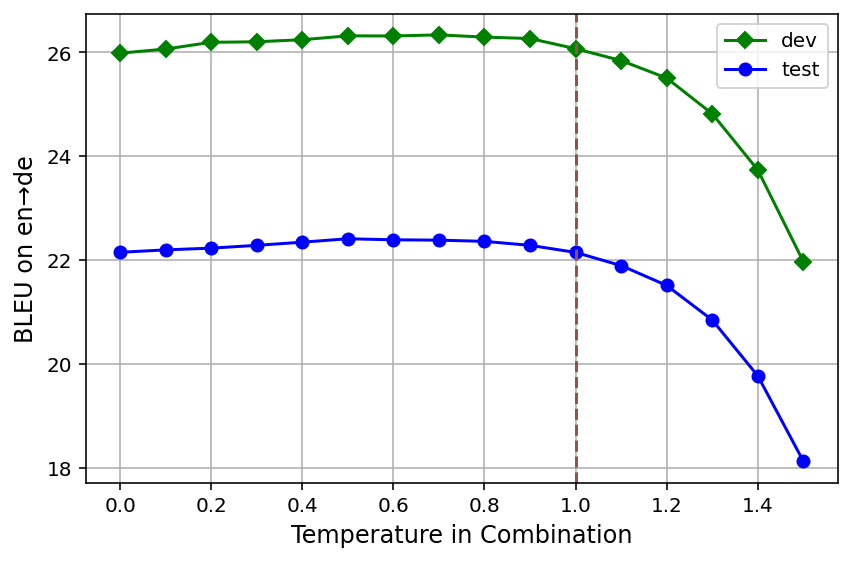}
  \end{minipage}
  \hfill
  \begin{minipage}{0.495\textwidth}
    \centering
    \includegraphics[width=\textwidth]{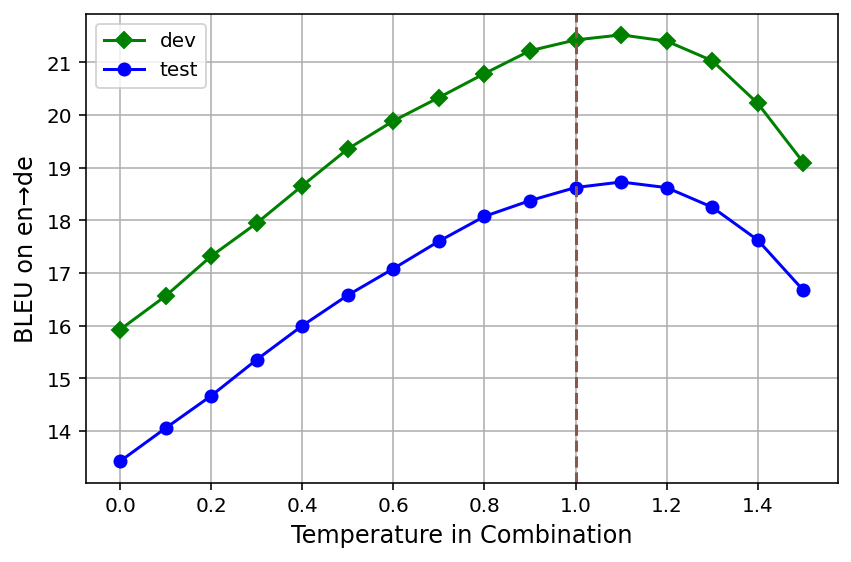}
  \end{minipage}
  \caption{Effect of changing the temperature value in LegoSLM on the CoVoST 2 \texttt{en$\rightarrow$de} speech translation task. For each experimental configuration, the speech encoder and LLM are trained under different setups and combined in a zero-shot fashion. Left: USM-CTC (public) + Gemma (multi). Right: USM-CTC (lbs) + Gemma (multi). }
  \label{ast_temp}
\end{figure}

Figure \ref{ast_temp} depicts the influence of $\tau$ on speech translation tasks where we can observe a similar trend that adjusting the temperature value results in improved BLEU scores on the CoVoST 2 test set.

Table \ref{tab:case} demonstrates the influence of the temperature value with two examples from the MLS\_en test set. For the first example, the optimal $\tau$ falls around 0.4 to 0.8, and for the second example, the optimal $\tau$ is in the range of 0.2 to 0.9. When $\tau$ is set to 0.5, fewer recognition errors can be observed compared to using $\tau$ equal to 1.0. By increasing $\tau$ to 1.5, the speech embeddings become less informative to LLM, leading to more erroneous hypotheses. In this case, the LLM leverages more of the stored world knowledge to complete the sentence. For instance, the second example generates \textit{coffee} in the decoding hypothesis at $\tau=1.5$, which is semantically close to \textit{beef tea} from the ASR reference. In the extreme case of $1e4$, the speech embeddings contain no valid information about the input utterance, causing the LLM to hallucinate and output repeating \textit{the united states of america} in the output.

\begin{table}[!htbp]
    \centering
    \small
    \begin{tabularx}{\linewidth}{l|l|X}
    \toprule
    Source & $\tau$ & Transcription \\
    \midrule
        \multirow{2}*{ASR-REF} & - & so among primitive men the weakest and stupidest went to the wall while the toughest and shrewdest those who were best fitted to cope with their circumstances but not the best in another way survived \\
    \midrule
       \multirow{12}*{\rotatebox{0}{LegoSLM}} & 1e-4 & so among primitive men the {\color{red}\textit{weaker}} and {\color{red}\textit{stupid}} went to {\color{red}\textit{*** war}} while {\color{red}\textit{were tough}} and {\color{red}\textit{shrewders}} those who were best {\color{red}\textit{pitted}} to cope with their circumstances but not the best in another way survived \\
       \cmidrule{2-3}
       & 0.5 & so among primitive men the weakest and stupidest went to {\color{red}\textit{*** war}} while the {\color{red}\textit{tough}} and {\color{red}\textit{shrewder}} those who were best fitted to cope with their circumstances but not the best in another way survived \\
       \cmidrule{2-3}
       & 1.0 & so among primitive men the {\color{red}\textit{weak}} and {\color{red}\textit{stupid ones}} went to {\color{red}\textit{*** war}} while the {\color{red}\textit{sagacious}} and {\color{red}\textit{shrewd ones}} those who were best fitted to cope with their circumstances but not the best in another way survived \\
       \cmidrule{2-3}
       & 1.5 & \textit{{\color{red}some of the most}} primitive men {\color{red}\textit{were}} the {\color{red}\textit{most weak}} and {\color{red}\textit{stupid of us}} went {\color{red}\textit{through our ordeal}} while the {\color{red}\textit{more sophisticated}} and {\color{red}\textit{shrewder ones}} those who were best fitted to cope with {\color{red}\textit{the new}} circumstances {\color{red}\textit{got out of}} the best {\color{red}\textit{and most enduring}} way {\color{red}\textit{survived}} \\
       \cmidrule{2-3}
       & 1e4 & \textit{\color{red} and the united states of america the united states of america the united states of america the united states of america the united states of america the united states of america ...} \\
    \midrule
    \midrule
    \multirow{2}*{ASR-REF} & - & at last however the beef tea was ready and valerie poured it into a cup which she stood in a bowl of cold water to cool it and then she hurried up with it to the child's room \\
    \midrule
       \multirow{12}*{\rotatebox{0}{LegoSLM}} & 1e-4 & at last however the \textit{\color{red}victorytea ***} was ready and \textit{\color{red}val} poured it into a cup which she stood in a bowl of cold water to cool it and then she hurried up with it to the \textit{\color{red}taos} room \\
       \cmidrule{2-3}
       & 0.5 & at last however the \textit{\color{red}victory} tea was ready and valeria poured it into a cup which she stood in a bowl of cold water to cool it and then she hurried up with it to the \textit{\color{red}taffrail} room \\
       \cmidrule{2-3}
       & 1.0 & at last however the \textit{\color{red}victory} tea was ready and valeria poured it into a cup which she stood in a bowl of cold water to cool and then she hurried up with it to the \textit{\color{red}tars tarkas} room \\
       \cmidrule{2-3}
       & 1.5 & at last however the \textit{\color{red}**** coffee} was ready and valeria poured it into a cup which she stood in a bowl of cold water to cool \textit{\color{red}**} and then she hurried \textit{\color{red}upstairs **** **} to the \textit{\color{red}tchinovniks} room \\
       \cmidrule{2-3}
       & 1e4 & \textit{\color{red} and the united states of america the united states of america the united states of america the united states of america the united states of america the united states of america ...} \\
    \bottomrule
    \end{tabularx}
    \caption{A case analysis of LegoSLM for ASR performance on the MLS\_en test set. The USM-CTC model is trained on MLS en-us data and Gemma is trained on LibriSpeech. Recognition errors are highlighted in \textit{\color{red}red color}.}
    \label{tab:case}
\end{table}

\end{document}